# Gaussian Mixture Model for Handwritten Script Identification

Mallikarjun Hangarge*

*Karnatak Arts,Science and Commerce College, Bidar 585401, Karnataka, India.

**Abstract**

This paper presents a Gaussian Mixture Model (GMM) to identify the script of handwritten words of Roman, Devanagari, Kannada and Telugu scripts. It emphasizes the significance of directional energies for identification of script of the word. It is robust to varied image sizes and different styles of writing. A GMM is modeled using a set of six novel features derived from directional energy distributions of the underlying image. The standard deviation of directional energy distributions are computed by decomposing an image matrix into right and left diagonals. Furthermore, deviation of horizontal and vertical distributions of energies is also built-in to GMM. A dataset of 400 images out of 800 (200 of each script) are used for training GMM and the remaining is for testing. An exhaustive experimentation is carried out at bi-script, tri-script and multi-script level and achieved script identification accuracies in percentage as 98.7, 98.16 and 96.91 respectively.

*Keywords: Gaussian Mixture Model; Handwritten Document; Script Identification;Document Image Analysis;*

## 1. Introduction

Script is a set of symbols and rules used to express or convey the information in a graphic form. It is independent of any specific language. Thus, different languages may use the same script. For example, Sanskrit, Marathi, and Hindi use the Devanagari script. Due to the diversity of Indian scripts and languages, a number of real time documents, which come across in day-to-day life are in multi-script/multi-lingual in nature. We can see the number of commercial and official documents containing local and Roman scripts. It implies that the existence of bi-script, tri-script and multi-script documents are common in India. To read such documents a robust multi-script OCR is essential. Multi-script OCR needs an automatic preprocessing model to identify the different script zones in a single document and then to pass it to a script specific OCR engine. So, the script identification and separation is a dynamic research problem of multi-lingual document image analysis. It has number of applications such as automatic text retrieval, sorting of text documents, enhancing the accuracy of OCR, facilitates indexing and multilingual document archival. Besides, it is essential to understand the shape of the characters of Indic scripts, before developing a multi-script OCR. The visual insight of shape of characters of each script shows that the characters of Indic script have horizontal, vertical, w-formation, inverted w-formation, holes and circular shape structures. These perceptions indicate that the shape of characters can be the composition of directional segments. Therefore, this paper makes use of statistics of directional energy distributions to characterize the shape of characters of scripts and demonstrates its significance in discriminating four major Indic scripts at word level.

### 1.1. Problem Overview

Handwritten script identification problem can be attempted mainly in two ways, namely top to bottom and bottom to top approach. In case of top to bottom approach, the problem can be viewed in three levels, specifically block, line and word level. In bottom to top approach, the task can be attempted at word, line and then block level. However, for Indian documents, word level script identification is the decisive solution and which enables us to read multi-lingual/multi-script documents. In fact, literature review reflects that most of the authors have dealt with this problem either at block level or line level or word level, but none in combination of these. In addition, the features

---

used in different algorithms (for example [2, 3]) are not generalized in nature. It means that the feature set used in discriminating text blocks is not efficient in discriminating text lines. Inline, the feature set employed for discriminating text blocks and text lines will not exhibit its efficacy in classifying the text words. Besides, a feature set used for classification of Kannada and Roman scripts are not efficient in classification of Malayalam and Roman scripts. This is because of the shape of the characters of one script differs from another. With these observations, attempting the problem at word level instead of attempting it at paragraph or line level is the interesting choice of study with respect to Indian documents.

*1.2. Previous Work*

It can be seen that almost script identification work reported in the literature is pertained to the printed text. However, less attention has been given to handwritten script identification. The handwritten documents of complex layouts, having variation in writing style, slanted characters, touched lines, different size characters, and unconstrained gaps between words etc., made the problem more challenging.

A detailed review on offline handwritten script identification can be found in [11] and complete survey on script identification can had it from [13]. For automatic recognition of script of the handwritten text, the various frequency domain features were proposed in the literature such as Discrete Cosine Transform (DCT), Gabor filters and Wavelets Transforms (WT) [1-3]. Besides, structural features like global, local, background and foreground, holistic, water reservoirs and relative Y-centroid, relative X-centroid, number of white holes, were proposed in [4-9]. These structural features are script dependent; however frequency domain features are not. But, frequency domain features are not proficient to work with smaller size images, for instance word images. It means that they work well for text blocks of larger size. Further, frequency domain features are not efficiently used to capture the directional energy distributions of characters of a script which plays a vital rule in shape analysis. For example, traditional DCT will emphasize on horizontal and vertical frequency responses and de-emphasizes minute edge energy frequency responses [12].

Based on these quick observations made in the aforementioned paragraph, a state-of-the-art technique is proposed to overcome the limitations by exploiting the directional energy distribution properties of the underlying image.

*1.3. Structure of the paper*

The paper is organized as follows. In detail explanation of the proposed method is given in Section 2. It mainly includes features extraction in Section 2.1 and classification in Section 2.2. Experimental results and a comprehensive analysis of results are reported in Section 3. We conclude our paper in Section 4.

## 2. Proposed Approach

The main aim is to identify the script of the word. To handle it, words are first extracted from the scanned document. For each extracted word, standard deviations of directional energy distributions are computed by decomposing an image matrix into right and left diagonals. Then these are integrated to form a single feature vector of dimension Nx1. Based on these features a Gaussian Mixture model has been built for each script to identify the script of the word. For better understanding, we refer readers to Fig. 1 where it provides a schematic work-flow of the system. The details of preprocessing technique used here may have it from [5].

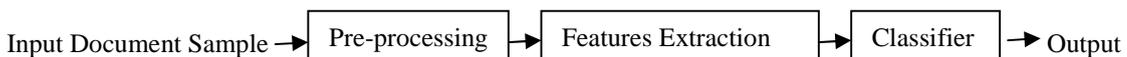

Fig. 1: A screen-shot of an overall work-flow of the system.

## 2.1. Feature Extraction

The process of feature extraction is established by converting each input image matrix into a square matrix by appending zeros in case of non square matrix. Let A be the square matrix of input image of size NXN. Let the principal diagonal of A be μ. Let    and    be N-2 upper and lower diagonals of A respectively. Then the computation of six features and are denoted by f1,…,f6, is discussed below. Firstly, extract a principal diagonal μ of A and compute its standard deviation $\sigma_1$ using

$$\sigma_1 = \left(\frac{1}{n-1}\sum_{i=0}^{n}(\mu_i - \overline{\mu})^2\right)^{1/2}, \tag{1}$$

where n is the number of elements in μ. The standard deviation $\sigma_1$ is a scalar value. Secondly, extract N-2 upper diagonals and compute their standard deviations using

$$X(\sigma_k) = \left(\frac{1}{n-1}\sum_{i=0}^{n}(\beta_{ik} - \overline{\beta})^2\right)^{1/2}, \tag{2}$$

where i=1,…, n and K=1,…,N-2. X is a coloum vector of size N-2X1.

Then, by appending the value of $\sigma_1$ and a zero into X, we get first feature f1 of dimension Nx1. Similarly, extract N-2 lower diagonals and compute their standard deviations using

$$Y(\sigma_m) = \left(\frac{1}{n-1}\sum_{i=0}^{n}(\alpha_{im} - \overline{\alpha})^2\right)^{1/2}, \tag{3}$$

where  m=1,….,N-2 and Y is a coloum vector of size N-2X1. By appending two zeros into Y, we get second feature f2 of dimension Nx1. In this way, features f3 and f4 are computed by filliping the input matrix A. The filliped matrix is denoted by $A^f$ and upper, lower and principal diagonals are by $\cdot^f$, $\cdot^f$ and $\mu^f$ respectively. Finally, standard deviations of horizontal and vertical energies of A are computed to obtain features f5 and f6 respectively.

## 2.2. Classification

Mainly, Gaussian Mixture Model (GMM) based script identification is a script classifier. Each script is modeled by GMM and script classification is carried out according to the likelihood score calculated by GMM against a given feature vector of unknown script. The Gaussian mixture model is a weighted sum of multivariate Gaussian mixture components. It forms the probability density function for a given set of feature vector and it is defined as

$$p(\vec{y}|\lambda) = \sum_{i=1}^{M} w_i\, b_i(\vec{y}) \tag{4}$$

where $w_i$ is the weight matrix, $\vec{y}$ is N dimensional feature vector, i is the mixture component index and $b_i(\vec{y})$ is an N-variate Gaussian density function. The complete GMM is defined by

$$\lambda = \{w(i), \mu(i), \Sigma i\} \tag{5}$$

The GMM parameters, $\lambda$ is computed by an iterative method using expectation maximization (EM) algorithm. Using this model, given a word of unknown script X= { $\vec{y}_1, \vec{y}_2, \vec{y}_3, \vec{y}_4, \ldots \vec{y}_T$ } is classified based on the average log likelihood score and is given by

$$\bar{L} = \arg\max_{i<i<L} \frac{1}{T}\sum_{t=1}^{T} \log\, p(\vec{y_t}|\lambda_t) \tag{6}$$

where $\lambda\_t$ is GMM of script i and L be the number of scripts the system could identify.

## 3. Experiments

*3.1. Dataset*

We have created a dataset of 800 word images of four scripts due to non availability of a benchmark dataset for Indic scripts. Ten handwritten pages of each script (Roman, Devnagari, Kannada and Telugu) written by 40 writers were collected and scanned at 300 dpi. The scanned documents are binarized using Otsu method [10]. Then words are segmented using the algorithm employed in [5]. A sample dataset of four scripts is shown in Fig 2.

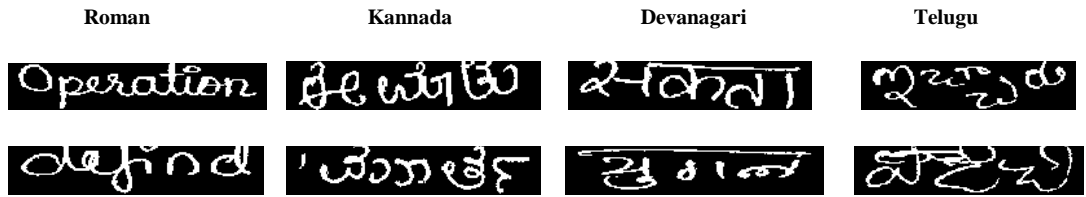

Fig. 2. Sample dataset of words of four scripts; First column represents the Roman script words; Second column represent words of Kannada script; Third column shows words of Devnagari script and Fourth column is of words of Telugu script.

*3.2. Results and Discussion*

An exhaustive experimentation has been carried out on a dataset of 800 word images. A Gaussian mixture model is designed for each script. It is trained with 400 samples out of 800 (100 samples of each script) and the remaining is used for testing. An experimental setup has been made in three phases, namely bi-script, tri-script and multi-script. In case of bi-script, Roman-Devnagari, Roman-Kannada and Roman-Telugu combinations are considered for experimentation. An average percentage of bi-script classification accuracy is 98.7% with GMM of order 128 and is reported in Table 1. The tri-script classification problem involves the combinations of Roman-Devnagari-Kannada and Roman-Devnagari-Telugu. In this case, average tri-script classification accuracy is 98.16% and it is shown in Table 2. The problem of multi-script identification is the classification problem of Roman, Devnagari, Kannada and Telugu scripts. The multi-script classification accuracies are reported in Table 3. Besides, experimentations are extended on multi-script classification to observe the performance of GMM with different orders of mixtures. And their respective results are illustrated in Fig. 3. These results indicate that the performance of the GMM is dependent on the number of mixture components considered for classification. The uniqueness of the algorithm is its robustness against variation in image size, size of characters, character slants, unconstrained gaps between the words, writing style. The experimental results exhibits that the performance of the directional energy features are significant in classifying different handwritten text words as compared with the results of [1, 2, 3]. Meanwhile, the decrease in classification accuracy is noticed in case of multi-script classification when scripts are increased and an investigation is on to overcome this problem.

Table 1. Average Percentage of Bi-script Identification accuracy with GMM of order 128.

| Scripts | Devnagari | Kannada | Telugu | Average % of Recognition |
|---------|-----------|---------|--------|--------------------------|
| Roman   | 97.0      | 100.0   | 99.0   | 98.7                     |

Table 2. Average Percentage of Tri-script Identification accuracy with GMM of order 128.

| Scripts→ | Roman | Kannada | Devnagari | Average % of Recognition | Roman | Telugu | Devnagari | Average % of Recognition |
|---|---|---|---|---|---|---|---|---|
| % of Recognition | 97.00 | 97.00 | 99.00 | 97.66 | 98.00 | 98.00 | 100.00 | 98.66 |

Table 3. Average Percentage of Multi-script Identification accuracy with GMM of order 128.

| Scripts→ | Roman | Kannada | Devnagari | Telugu | Average % of Recognition |
|---|---|---|---|---|---|
| % of Recognition | 96 | 95 | 97.67 | 99 | 96.91 |

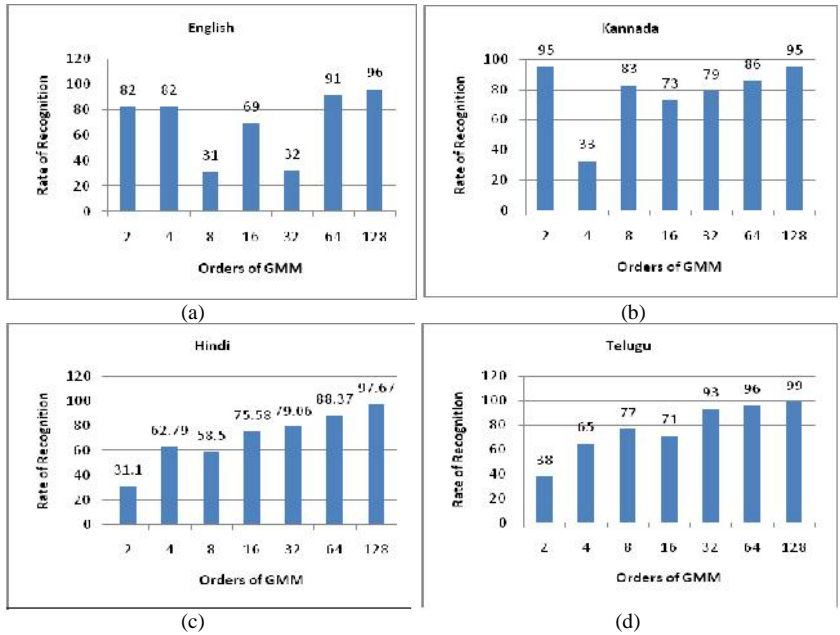

Fig. 3. Illustrates the percentage of script identification accuracies in case of multi-scripts: (a) Percentage of Roman script identification accuracies with different orders of GMM; (b) Percentage of Kannada script identification accuracies with different orders of GMM; (c) Percentage of Devanagari script identification accuracies with different orders of GMM; (d) Percentage of Telugu script identification accuracies with different orders of GMM

## 4. Conclusion and Future Work

Any shape can be quantified as the combination of directional primitives like line segments that holds significant information about the shape. The shape of characters of Indic script gives us a clue for computing directional energies. This work emphasizes the use of the directional energy features in classification of the proposed scripts. Experimental results confirm the potentiality of the proposed features. Dimensionality and computing complexities of the features are near to the ground. Features are simple in nature and invariant to image size, font size, noise, slants of characters, broken characters etc. In future, we extend this work to evaluate the performance of the proposed method against DCT, DWT, Gabor filters and steerable filters on inclusion of more number of scripts.

**Acknowledgement**

This work is carried out under the UGC sponsored minor research project (ref: MRP(S):661/09-10/KAGU013/UGCSWRO dated, 30/11/2009).